\documentclass[twoside,11pt]{article}

\usepackage{jmlr2e}
\usepackage[numbers, square, comma, sort]{natbib}

\usepackage{subcaption}
\usepackage{amssymb}
\usepackage{amsmath}
\usepackage{mathtools}
\usepackage{nicefrac}
\usepackage{algorithm}
\usepackage{svg}
\usepackage[noend]{algpseudocode}
\usepackage{setspace}
\usepackage{sectsty}

\allowdisplaybreaks
\sectionfont{\fontsize{16}{15}\selectfont}
\let\originalparagraph\paragraph
\renewcommand{\paragraph}[1]{\vspace{-1.5em}\originalparagraph{#1}}

\ShortHeadings{Distributed Optimization using Heterogeneous Compute Systems}{Vineeth}
\firstpageno{1}

\begin{document}

\title{\LARGE Distributed Optimization using Heterogeneous Compute Systems}

\author{\name Vineeth S \email vineeths@iisc.ac.in \\
       \addr Division of Electrical, Electronics, and Computer Sciences\\
       Indian Institute of Science\\
       Bangalore, India}


\maketitle

\begin{abstract}
	Hardware compute power has been growing at an unprecedented rate in recent years. The utilization of such advancements plays a key role in producing better results in less time -- both in academia and industry. However, merging the existing hardware with the latest hardware within the same ecosystem poses a challenging task. One of the key challenges, in this case, is varying compute power. In this paper, we consider the training of deep neural networks on a distributed system of workers with varying compute power. A naive implementation of synchronous distributed training will result in the faster workers waiting for the slowest worker to complete processing. To mitigate this issue, we propose to dynamically adjust the data assigned for each worker during the training. We assign each worker a partition of total data proportional to its computing power. Our experiments show that dynamically adjusting the data partition helps to improve the utilization of the system and significantly reduces the time taken for training. Code is available at the repository: \url{https://github.com/vineeths96/Heterogeneous-Systems}.
\end{abstract}


\newpage
\section{Introduction}
Hardware computing power has been growing at a rapid rate in the past few years. Interestingly, GPU computing power has been growing at a rate greater than Moore's law \cite{hernandez2020measuring}. This growth has facilitated the training of large-scale models such as Google Switch Transformer (1.6 trillion parameters) \cite{fedus2021switch} and OpenAI GPT-3 Transformer (175 billion parameters) \cite{brown2020language}. Large-scale models are typically trained in distributed clusters with hundreds of GPUs to utilize the computing power of multiple workers. This paradigm of distributed training is known as data parallelism \cite{105555}\cite{124408}. In the data parallelism approach, the entire dataset is partitioned and assigned to workers. Workers then proceed to train on their assigned partition and synchronize their gradients with other workers at the end of every iteration. The worker gradients are averaged and then used for updating the model parameters. Data parallelism works well in practice and is widely used for distributed training nowadays. Distributed learning can be mathematically formulated as an optimization problem as follows:

\begin{align}
	\begin{split}
	\min_{\theta \in \mathbb{R}^{d}} f(\boldsymbol{\theta}) &= \frac{1}{M} \sum_{m=1}^{M} f_{m}(\boldsymbol{\theta}) \\
	f_{m}(\boldsymbol{\theta}) &= \mathop{\mathbb{E}}_{\mathbf{x}^{n} \sim \mathcal{D}^{m}} f(\boldsymbol{\theta}; \mathbf{x}^{n}) ,
	\end{split} 
\end{align}

where $\boldsymbol{\theta} \in \mathbb{R}^{n}$ denotes the model parameters to be learned, $M$ is the number of workers, $\mathcal{D}^{m}$ denotes the local data distribution at worker $m$, and $f(\boldsymbol{\theta}; \mathbf{x})$ is the loss function of the model on input $\mathbf{x}$. At each iteration, workers train a local copy of the model with a subset of the total data. We assume the standard optimization setting where every worker $m$ can locally observe an unbiased stochastic gradient $\mathbf{g}^{m}_{t}$, such that $ \mathbb{E}[\mathbf{g}^{m}_{t}] = \nabla f_{m}(\boldsymbol{\theta}_{t})$ at iteration $t$. Workers sample the local stochastic gradients $\mathbf{g}^{m}_{t}$ in parallel, and they synchronize them by averaging among the workers $\mathbf{g}_{t} = \frac{1}{M} \sum_{m=1}^{M} \mathbf{g}^{m}_{t}$. The model parameters are then iteratively updated using Stochastic Gradient Descent (SGD) or one of its variants with $\boldsymbol{\theta}_{t+1} = \boldsymbol{\theta}_{t}- \eta_{t} \mathbf{g}_{t}$, where $\eta_{t}$ is the learning rate at iteration $t$.

Data parallelism is efficient as long as all the workers have similar compute power. However, when the workers have different compute powers, the slower workers will straggle the faster workers: the faster workers will have to wait for the slower works to complete their processing and synchronize. This results in the underutilization of the faster worker resources. Straggler mitigation and its application to machine learning is a well researched domain \cite{10.1145/2408776.2408794}\cite{10.5555/2482626.2482645}\cite{inproceedings}\cite{180560}. Most of the existing work on straggler mitigation assumes that workers have (almost) identical compute power -- with the runtime variability arising due to alternate factors. The key difference between the existing research and the problem addressed in this paper is that we have workers with different computing power. The workers with lesser computing power will be straggling the workers with higher computing power always. In other words, we have \textit{deterministic stragglers}. If the tools and techniques developed for straggler mitigation for workers with identical computing power are used in this setting, the results will be suboptimal. Those methods will be biased towards the workers with lower computing power since they are not designed for this scenario.

We consider a distributed system of workers with varying compute powers. We call these types of systems as \textit{heterogeneous systems}. Heterogeneous systems can arise in many scenarios in practice. For example, when integrating the latest hardware with existing hardware in the same ecosystem or in federated learning where the devices have different computational capacities. The AI Index Report from Stanford \cite{aiindex} notes that the AI compute power has approximately doubled every 3.4 months since 2012. At this rate of new hardware releases, it would be economically infeasible to upgrade the hardware ecosystem consistently. Naively integrating the existing hardware with the latest hardware will cause performance issues mentioned earlier. In federated learning, the compute power heterogeneity arises naturally as a result of the wide range of possible clients taking part in the training \cite{mcmahan2017communicationefficient}. Despite this, there is very limited research in this direction which makes this pursuit challenging and interesting.

\paragraph{Contributions} We propose dynamical partitioning of the dataset to workers at every epoch. We partition the dataset in proportion to the computing power of the workers. By adjusting the data partition to the workers, we directly control the workload on the workers. We assign the partitions, and hence the workloads, such that the time taken to process the data partition is almost uniform across the workers. We empirically evaluate the performance of the dynamic partitioning by training deep neural networks on the CIFAR10 dataset \cite{article}. We examine the performance of training ResNet50 (computation-heavy) model and VGG16 (computation-light) model with and without the dynamic partitioning algorithms. We implement the heterogeneous framework in PyTorch \cite{paszke2019pytorch}.

\section{Related Work}
Straggler mitigation is a well-studied topic in research. Recent years have also witnessed the development of straggler mitigation techniques specifically for machine learning such as data encoding \cite{karakus2018straggler}, task replication \cite{10.1145/2847220.2847223}, and gradient coding \cite{pmlr-v70-tandon17a}. Most of the prior work in straggler mitigation treats the workers to have identical compute power with system noise that varies the worker performance. However, these assumptions do not fit in a heterogeneous cluster, as discussed earlier. Another line of recent research in straggler mitigation is choosing the fastest $k$ worker updates and ignoring the straggler updates \cite{chen2017revisiting}\cite{dutta2018slow}. This method would be detrimental in a heterogeneous computing environment, as the workers with the lower compute power will always be ignored.

\section{Preliminaries}
We consider minimizing an objective function with synchronous stochastic gradient descent (Sync-SGD) on workers in a heterogeneous cluster. We assume there are $M$ workers. In a typical Sync-SGD setting, the total dataset $\mathcal{D}$ will be divided uniformly across workers as $\mathcal{D}_{1}$, $\mathcal{D}_{2}$, $\dots$, $\mathcal{D}_{M}$. Each partition $\mathcal{D}_{i}$ will roughly have $\frac{\lvert \mathcal{D} \rvert}{M}$ data samples, where $\lvert \mathcal{D} \rvert$ denotes the total size of the dataset. At each iteration, worker $m$ samples a mini-batch from its partition $\mathcal{D}_{m}$, trains the model, and calculates the gradient for that mini-batch. Workers then exchange their gradients with other workers via a centralized framework (e.g. parameter server) or decentralized framework (e.g. all-reduce). This exchange of gradients is the synchronization step that becomes a bottleneck in the presence of stragglers. Once the gradients are exchanged, they are averaged, and the model is updated. 

If all the workers have identical compute power, the time taken for processing a mini-batch will be equal in expectation. On the other hand, if the computing power of workers is different, each worker will take a different amount of time to process a mini-batch. Assume random variable $X_{m}$ to be the time taken to process a mini-batch by worker $m$. Then, the time taken for every iteration for Sync-SGD would be $X_{(M)}$, where $X_{(k)}$ is the $k^{th}$ order statistic of random variables $X_{1}$, $X_{2}$, $\dots$, $X_{M}$. We wish to minimize the expected value of $X_{(M)}$ by dynamically adjusting the data partitions assigned to the workers.

\section{Algorithms}
In this section, we present the proposed algorithms for efficient utilization of heterogeneous systems for machine learning -- Dynamic Partitioning and Enhanced Dynamic Partitioning. The key idea in the algorithms is to adjust the data partition $\mathcal{D}_{m}$ for worker $m$ such that time taken for processing a mini-batch $X_{m}$ is almost uniform and $X_{(M)}$ is minimized. We achieve this by assigning data partition $\mathcal{D}_{m}$ in proportion to the computing power of worker $m$. However, in practice, we may not know the computing power of individual workers in a distributed group beforehand. We estimate the computing power of a worker online by calculating the number of stochastic gradients calculated per second by the worker. Intuitively, the worker with higher compute power will calculate more stochastic gradients per second, and the worker with lower compute power will calculate less stochastic gradients per second.

The compute power estimate of worker $m$ which processed $\mathcal{D}_{m}$ data partition in $X_{m}$ time,
\begin{equation}
	CP_{m} = \frac{\lvert \mathcal{D}_{m} \rvert}{X_{m}}.
\end{equation}

Hence, the approximate time taken for processing $\mathcal{D}^{\prime}_{m}$ data partition is given by,
\begin{equation}
	X^{\prime}_{m} = \frac{\lvert \mathcal{D}^{\prime}_{m} \rvert}{CP_{m}}.
\end{equation}
We aim to make the $X^{\prime}_{m}$s uniform across all workers. In order to achieve the goal, we can that see we need to partition the dataset into $\mathcal{D}^{\prime}_{m}$s in proportion to $CP_{m}$s.

\subsection{Dynamic Partitioning}\label{sec:dyn_par}
The Dynamic Partitioning algorithm proceeds as follows. At the beginning of the training (for the first epoch), the data is partitioned evenly across the workers. Workers train on their assigned partition by sampling mini-batches from their partition. The mean time taken to process a mini-batch of data is tracked for every worker over an epoch. Using this, we calculate the estimate for computing power, $CP$, for the workers. We can use the information from just the previous epoch or use the information from a history of epochs. Using the time from just the previous epoch will allow the algorithm to adapt to dynamic changes in computing power rapidly. These dynamic changes can arise due to factors such as workload, temperature, etc. On the other hand, using the time from a history of epochs will be resilient to such dynamic changes, which will provide a better estimate of hardware computing power. From the second epoch onwards, we distribute the data in proportion to the estimated $CP$s and iterate over the procedure. We estimate computing power, $CP$, every epoch using information from the previous epochs. Algorithm \ref{alg:dyn_par} shows the Dynamic Partitioning algorithm where $\mathbf{X}_{t}$ and $\boldsymbol{\mathcal{D}}_{t}$ are vectors that represent the mean batch process times and dataset partitions for the workers, respectively.

\begin{algorithm}
	\caption{Dynamic Partitioning}\label{alg:dyn_par}\onehalfspacing
	\begin{algorithmic}[1]
		\State \textbf{Input}: Epoch $t$, Mean Process Times $\mathbf{X}_{t}$, Dataset partitions $\boldsymbol{\mathcal{D}}_{t}$, Dataset $\mathcal{D}$, and number of workers $M$
		\Procedure{DynamicPartitioning}{$\mathbf{X}_{t}$, $\boldsymbol{\mathcal{D}}_{t}$, $\mathcal{D}$, t} 
		\If {$t == 0$} 
		\ForAll{worker $m$}:
		\State Reset $CP_{m} = \nicefrac{1}{M}$ \Comment{Initialize partitions evenly} 
		\EndFor
		
		\Else
		\ForAll{worker $m$}:
		\State Calculate $CP_{m} = \frac{\lvert \mathcal{D}_{m} \rvert}{X_{m}}$ \Comment{Estimate compute power}		
		\EndFor
		
		\State Calculate $CP_{m} = \frac{CP_{m}}{\sum_{j} CP_{j}}$ \Comment{Normalize compute powers}
		\State Partition dataset $\mathcal{D}$ into $\boldsymbol{\mathcal{D}}_{t+1}$ such that 
		
		\hspace{10em}$\lvert \boldsymbol{\mathcal{D}}^{m}_{t+1} \rvert = CP_{m} \times \lvert \mathcal{D} \rvert$ 
		\EndIf
		\EndProcedure
	\end{algorithmic}
\end{algorithm}

\subsection{Enhanced Dynamic Partitioning}\label{sec:dyn_par_e}
The Enhanced Dynamic Partitioning algorithm proceeds the same way as Dynamic Partitioning algorithm, except for a ``heuristic enhancement" step. Intuitively, the training process should give more preference to the worker that processed a larger data partition. To incorporate this enhancement, we weigh the gradients $\mathbf{g}^{m}_{t}$ by the normalized compute power estimate of the respective workers, $CP_{m}$. We do this weighting at every iteration before the gradient communication step.


\section{Experiments}
We implement our code in PyTorch \cite{paszke2019pytorch} using NVIDIA NCCL communication library. We validate our approach with the image classification on the CIFAR10 dataset \cite{article}. We examine the performance of training ResNet50 (computation-heavy) model and VGG16 (computation-light) model, with and without the dynamic partitioning methods. We train for 150 epochs with a batch size of 128 per worker (weak scaling). We use SGD optimizer with a momentum of $0.9$ and weight decay of $5 \times 10^{-4}$ in conjunction with Cosine Annealing LR \cite{loshchilov2017sgdr} scheduler. We repeat the experiments five times and plot the mean and standard deviation for all graphs.

We begin with an explanation of the notations used for the plot legends in this section. \textit{Sync-SGD} corresponds to the default gradient aggregation provided by PyTorch. \textit{DP-SGD} and \textit{EDP-SGD} corresponds to \textit{Dynamic Partitioning} (Section \ref{sec:dyn_par}) and \textit{Enhanced Dynamic Partitioning} (Section \ref{sec:dyn_par_e}) respectively. Unfortunately, we did not have access to a heterogeneous system at the time of writing this thesis. We artificially simulate heterogeneity by adding time delays to a subset of workers. We evaluate the algorithms for a low level of heterogeneity and a high level of heterogeneity.

\subsection{AWS Case study}
\begin{figure}[!h]
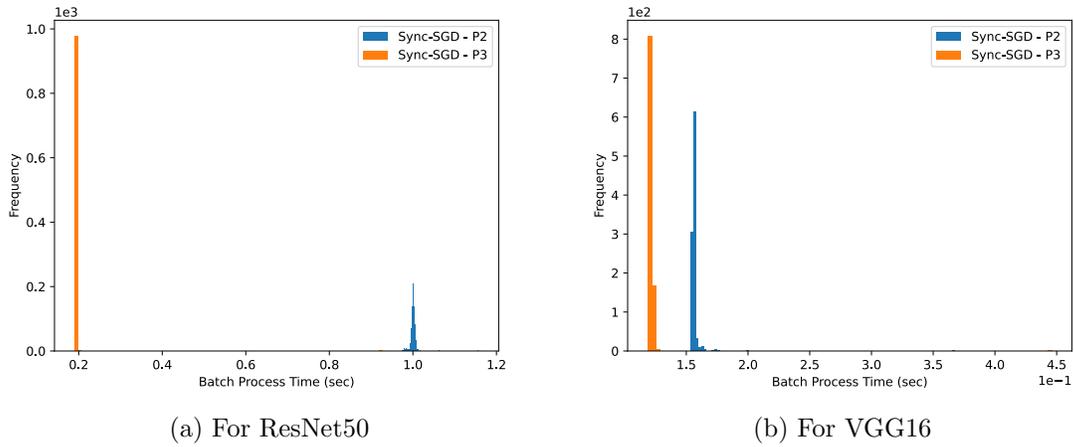

	\centering
	\begin{subfigure}{0.5\textwidth}
		\includesvg[width=\linewidth]{./imgs/AWS_Process_Times/AWS_process_times_histogram_ResNet50_NoneAllReducer.svg}
		\caption{For ResNet50}
		\label{fig:histograma}
	\end{subfigure}%
	\begin{subfigure}{0.5\textwidth}
		\includesvg[width=\linewidth]{./imgs/AWS_Process_Times/AWS_process_times_histogram_VGG16_NoneAllReducer.svg}
		\caption{For VGG16}
		\label{fig:histogramb}
	\end{subfigure}
	\caption{AWS Heterogeneous System: Batch Process Time Histogram}
	\label{fig:histogram}
\end{figure}

We first investigate how a practical heterogeneous system would look like by training on AWS instances. More specifically, we choose previous generation p2.xlarge (P2) and current generation p3.2xlarge instances (P3) and profile the time taken to process one mini-batch of data. Figures \ref{fig:histograma} and \ref{fig:histogramb} show the histogram of the time taken to process one mini-batch for the ResNet50 architecture and VGG16 architecture respectively. We can observe that the difference in the batch process times is larger for a compute-heavy model like ResNet50 and is smaller for a compute-light model like VGG16. We can also note that the batch process time for the cheaper and previous generation P2 worker has a higher variance than the latest generation P3 worker.

\begin{figure}[!h]
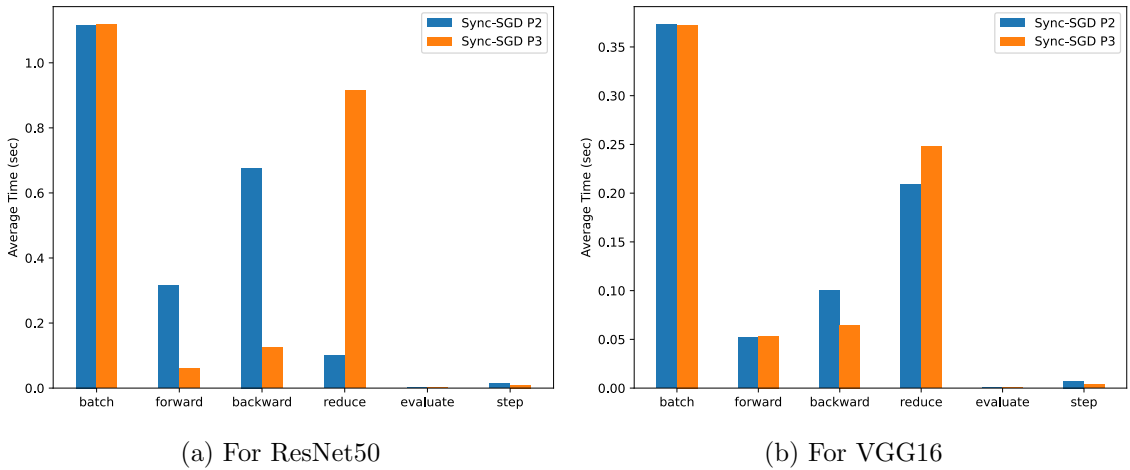

	\centering
	\begin{subfigure}{0.5\textwidth}
		\includesvg[width=\linewidth]{./imgs/AWS_Time_Breakdown/hetsys_time_breakdown_ResNet50.svg}
		\caption{For ResNet50}
		\label{fig:breakdowna}
	\end{subfigure}%
	\begin{subfigure}{0.5\textwidth}
		\includesvg[width=\linewidth]{./imgs/AWS_Time_Breakdown/hetsys_time_breakdown_VGG16.svg}
		\caption{For VGG16}
		\label{fig:breakdownb}
	\end{subfigure}
	\caption{AWS Heterogeneous System: Time Breakdown}
	\label{fig:breakdown}
\end{figure}

Figures \ref{fig:breakdowna} and \ref{fig:breakdownb} show the breakdown of time taken for various processes during training in this heterogeneous setup. We can observe that the faster P3 worker spends less time in computing the forward-backward passes and more time waiting in the communication. On the other hand, the slower P2 worker takes more time in computing and slows down the whole training process. This straggling effect is more pronounced in ResNet50 than VGG16 because of the respective computation complexity.

\subsection{Low Heterogeneity}
\begin{figure}[!h]
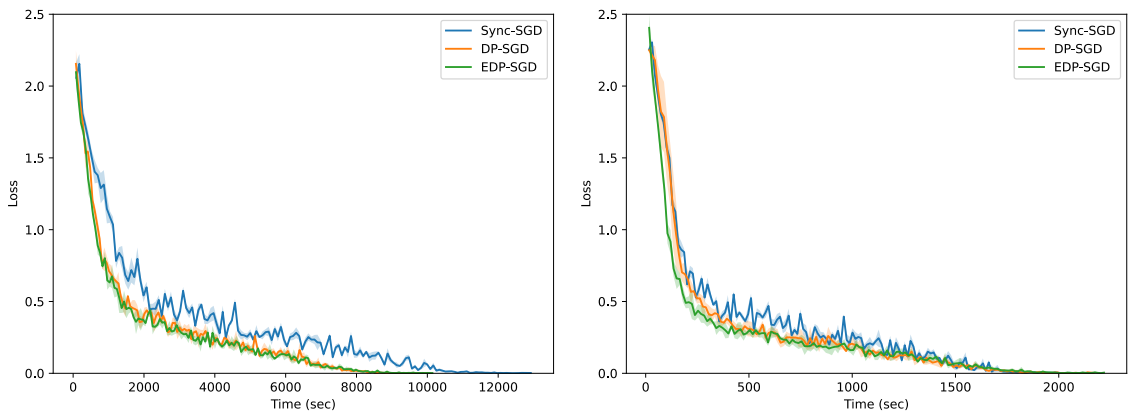

	\centering
	\begin{subfigure}{0.5\textwidth}
		\includesvg[width=\linewidth]{./imgs/Delay_1/1_loss_time_ResNet50.svg}
		\caption{For ResNet50}
		\label{fig:low_loss_timea}
	\end{subfigure}%
	\begin{subfigure}{0.5\textwidth}
		\includesvg[width=\linewidth]{./imgs/Delay_1/1_loss_time_VGG16.svg}
		\caption{For VGG16}
		\label{fig:low_loss_timeb}
	\end{subfigure}
	\caption{Low Heterogeneity: Loss vs Time}
	\label{fig:low_loss_time}
\end{figure}

\begin{figure}[!h]
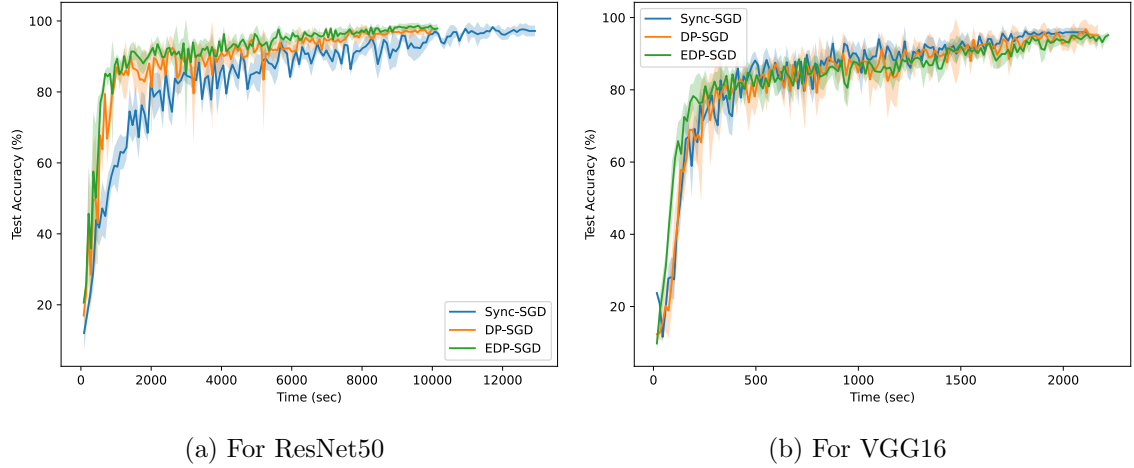

	\centering
	\begin{subfigure}{0.5\textwidth}
		\includesvg[width=\linewidth]{./imgs/Delay_1/1_top1_time_ResNet50.svg}
		\caption{For ResNet50}
		\label{fig:low_top1_timea}
	\end{subfigure}%
	\begin{subfigure}{0.5\textwidth}
		\includesvg[width=\linewidth]{./imgs/Delay_1/1_top1_time_VGG16.svg}
		\caption{For VGG16}
		\label{fig:low_top1_timeb}
	\end{subfigure}
	\caption{Low Heterogeneity: Accuracy vs Time}
	\label{fig:low_top1_time}
\end{figure}

In this section, we compare the performance of the algorithms with the native PyTorch aggregation for low heterogeneity. For simulating a low level of heterogeneity, we add delays of $0.1 sec$ for the ResNet50 architecture and $0.01 sec$ for the VGG16 architecture (roughly 25\% of the mean mini-batch process time). Figures \ref{fig:low_loss_time} and \ref{fig:low_top1_time} show the training loss and test accuracy respectively, over the time during training. 

From figures \ref{fig:low_loss_timea} and \ref{fig:low_top1_timea}, we can observe that dynamic partitioning performs better than the Sync-SGD for ResNet50 architecture. We can observe that the time taken for training is significantly lesser for our algorithms. From figure \ref{fig:low_loss_timeb} and \ref{fig:low_top1_timeb}, we can observe that dynamic partitioning performs slightly better than the Sync-SGD for VGG16 architecture.


However, we can also note that the training time is slightly more than Sync-SGD. We can explain this with the lower computation complexity of VGG16 architecture and the overhead of communication of partitioning information at every epoch. The proposed algorithms reduce the training time significantly for computation-heavy models. The enhanced dynamic partitioning performs marginally better than dynamic partitioning for ResNet50 architecture, as visible in the accuracy plots. 

\begin{figure}[!h]
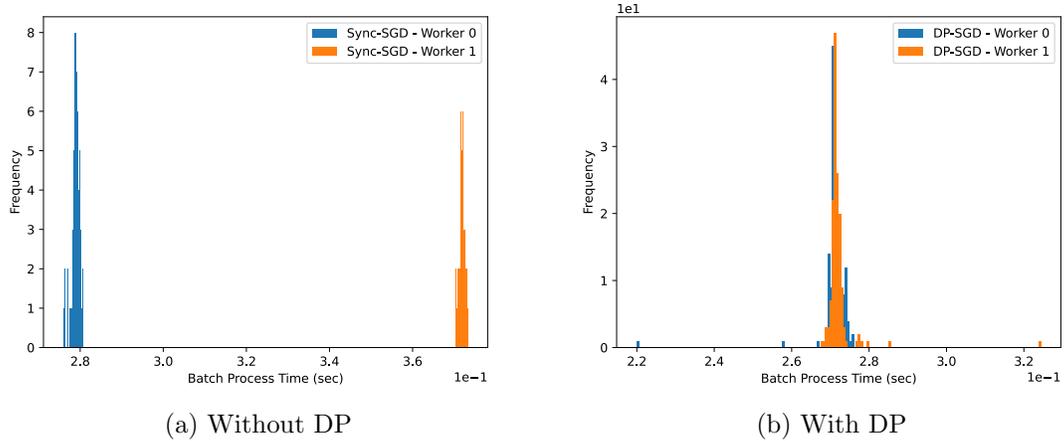

	\centering
	\begin{subfigure}{0.5\textwidth}
		\includesvg[width=\linewidth]{./imgs/Histograms_0.1/1_process_times_histogram_ResNet50_False_AllReduce_SGD.svg}
		\caption{Without DP}
		\label{fig:low_hist_bef_resnet}
	\end{subfigure}%
	\begin{subfigure}{0.5\textwidth}
		\includesvg[width=\linewidth]{./imgs/Histograms_0.1/1_process_times_histogram_ResNet50_True_AllReduce_SGD.svg}
		\caption{With DP}
		\label{fig:low_hist_aft_resnet}
	\end{subfigure}
	\caption{Low Heterogeneity: Batch Process Time Histogram for ResNet50}
	\label{fig:low_hist_resnet}
\end{figure}

\begin{figure}[!h]
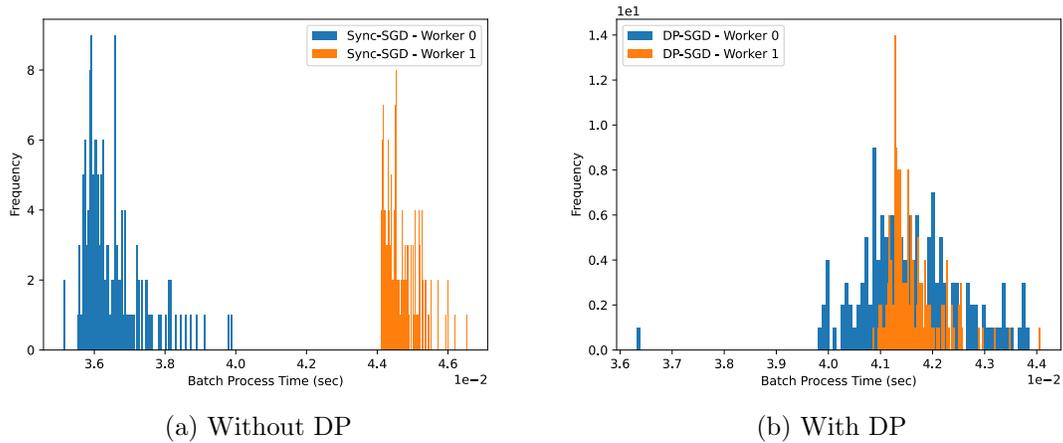

	\centering
	\begin{subfigure}{0.5\textwidth}
		\includesvg[width=\linewidth]{./imgs/Histograms_0.1/1_process_times_histogram_VGG16_False_AllReduce_SGD.svg}
		\caption{Without DP}
		\label{fig:low_hist_bef_vgg}
	\end{subfigure}%
	\begin{subfigure}{0.5\textwidth}
		\includesvg[width=\linewidth]{./imgs/Histograms_0.1/1_process_times_histogram_VGG16_True_AllReduce_SGD.svg}
		\caption{With DP}
		\label{fig:low_hist_aft_vgg}
	\end{subfigure}
	\caption{Low Heterogeneity: Batch Process Time Histogram for VGG16}
	\label{fig:low_hist_vgg}
\end{figure}

Figures \ref{fig:low_hist_resnet} and \ref{fig:low_hist_vgg} show the histograms for mini-batch processing time for ResNet50 architecture and VGG16 architecture, respectively. We can observe from the figures that dynamic partitioning helps to reduce the difference in the mini-batch process times of the workers. This will reduce the time spent by the faster worker waiting for the slower worker during the communication phase, effectively remedying straggling.

\subsection{High Heterogeneity}
In this section, we compare the performance of the algorithms with the native PyTorch aggregation for high heterogeneity. 


\begin{figure}[!h]
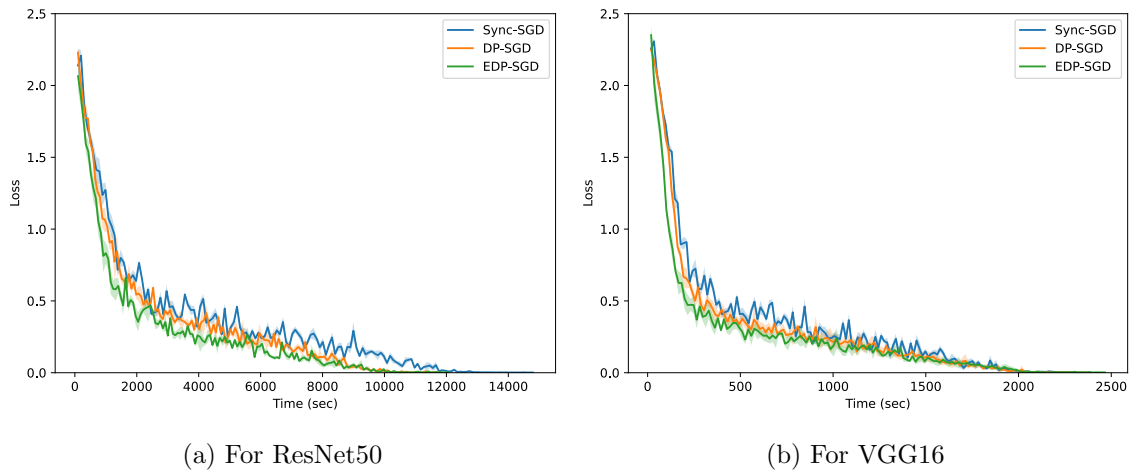

	\centering
	\begin{subfigure}{0.5\textwidth}
		\includesvg[width=\linewidth]{./imgs/Delay_2/2_loss_time_ResNet50.svg}
		\caption{For ResNet50}
		\label{fig:high_loss_timea}
	\end{subfigure}%
	\begin{subfigure}{0.5\textwidth}
		\includesvg[width=\linewidth]{./imgs/Delay_2/2_loss_time_VGG16.svg}
		\caption{For VGG16}
		\label{fig:high_loss_timeb}
	\end{subfigure}
	\caption{High Heterogeneity: Loss vs Time}
	\label{fig:high_loss_time}
\end{figure}


\begin{figure}[!h]
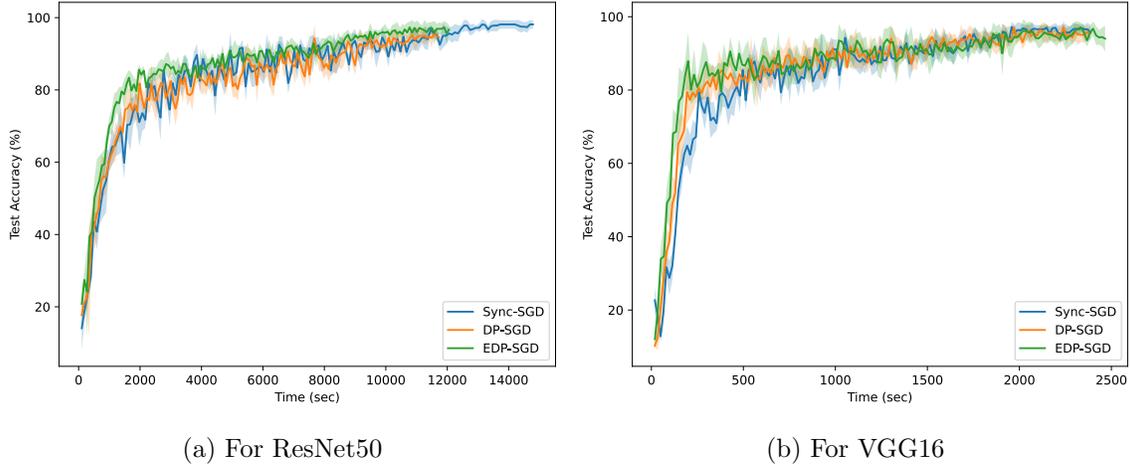

	\centering
	\begin{subfigure}{0.5\textwidth}
		\includesvg[width=\linewidth]{./imgs/Delay_2/2_top1_time_ResNet50.svg}
		\caption{For ResNet50}
		\label{fig:high_top1_timea}
	\end{subfigure}%
	\begin{subfigure}{0.5\textwidth}
		\includesvg[width=\linewidth]{./imgs/Delay_2/2_top1_time_VGG16.svg}
		\caption{For VGG16}
		\label{fig:high_top1_timeb}
	\end{subfigure}
	\caption{High Heterogeneity: Accuracy vs Time}
	\label{fig:high_top1_time}
\end{figure}

For simulating a high level of heterogeneity, we add delays of $0.2 sec$ for the ResNet50 architecture and $0.02 sec$ for the VGG16 architecture (roughly 50\% of the mean mini-batch process time). Figures \ref{fig:high_loss_time} and \ref{fig:high_top1_time} show the training loss and test accuracy respectively, over the time during training. From figures \ref{fig:high_loss_timea}, \ref{fig:high_top1_timea}, \ref{fig:high_loss_timeb} and \ref{fig:high_top1_timeb}, we can derive observations similar to those obtained for low heterogeneity. We can also observe that enhanced dynamic partitioning has a better training profile compared to dynamic partitioning. 

\begin{figure}[!h]
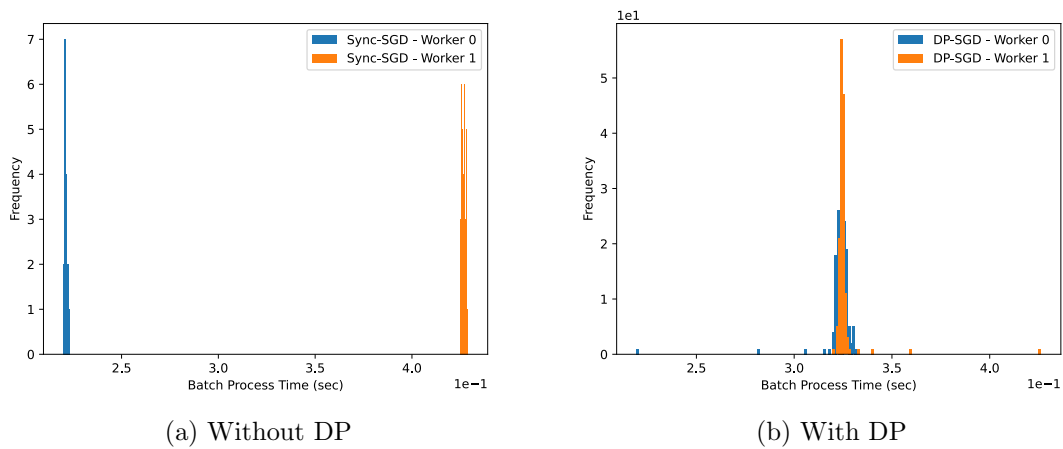

	\centering
	\begin{subfigure}{0.5\textwidth}
		\includesvg[width=\linewidth]{./imgs/Histograms_0.2/2_process_times_histogram_ResNet50_False_AllReduce_SGD.svg}
		\caption{Without DP}
		\label{fig:high_hist_bef_resnet}
	\end{subfigure}%
	\begin{subfigure}{0.5\textwidth}
		\includesvg[width=\linewidth]{./imgs/Histograms_0.2/2_process_times_histogram_ResNet50_True_AllReduce_SGD.svg}
		\caption{With DP}
		\label{fig:high_hist_aft_resnet}
	\end{subfigure}
	\caption{High Heterogeneity: Batch Process Time Histogram for ResNet50}
	\label{fig:high_hist_resnet}
\end{figure}

\begin{figure}[!h]
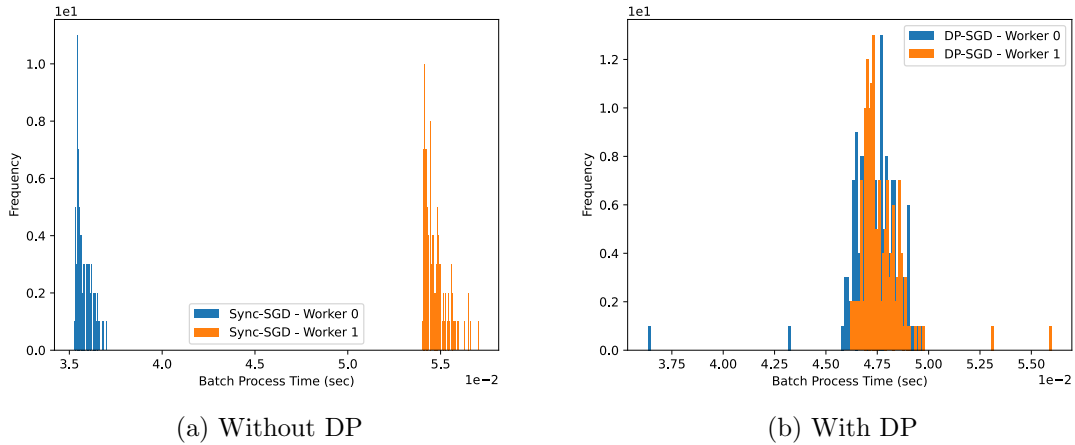

	\centering
	\begin{subfigure}{0.5\textwidth}
		\includesvg[width=\linewidth]{./imgs/Histograms_0.2/2_process_times_histogram_VGG16_False_AllReduce_SGD.svg}
		\caption{Without DP}
		\label{fig:high_hist_bef_vgg}
	\end{subfigure}%
	\begin{subfigure}{0.5\textwidth}
		\includesvg[width=\linewidth]{./imgs/Histograms_0.2/2_process_times_histogram_VGG16_True_AllReduce_SGD.svg}
		\caption{With DP}
		\label{fig:high_hist_aft_vgg}
	\end{subfigure}
	\caption{High Heterogeneity: Batch Process Time Histogram for VGG16}
	\label{fig:high_hist_vgg}
\end{figure}

Figures \ref{fig:high_hist_resnet} and \ref{fig:high_hist_vgg} show the histograms for mini-batch processing time for ResNet50 architecture and VGG16 architecture, respectively. We can observe from the figures that dynamic partitioning remedies straggling as seen earlier with low heterogeneity case.

\section{Concluding Remarks}
With rapid developments in the hardware industry, the computing power of devices introduced to the world keeps growing. This gives rise to heterogeneous systems, which can result in straggling during computation. In this paper, we presented algorithms for the efficient utilization of a heterogeneous system for distributed machine learning. We conduct extensive experiments on practical deep learning models and evaluate the performance of the proposed dynamic partitioning algorithms. We empirically show that dynamic partitioning helps to mitigate the straggling effect by adjusting the workload of the workers according to their computing power. The savings in time is significant when training compute-heavy models, such as ResNet50 model.

\newpage
\acks{The author would like to thank Himanshu Tyagi for suggesting the problem, useful discussions, and guidance throughout this work. This work has been supported by a research grant from the Robert Bosch Center for Cyber-Physical Systems (RBCCPS) at the Indian Institute of Science, Bangalore.}

\bibliographystyle{unsrtnat}
\bibliography{references}

\end{document}